\documentclass[11pt]{article}

\usepackage[final]{acl}

\usepackage{times}
\usepackage{latexsym}

\usepackage[T1]{fontenc}

\usepackage[utf8]{inputenc}

\usepackage{microtype}

\usepackage{inconsolata}

\usepackage{graphicx}

\usepackage{hyperref}
\usepackage{xurl}
\usepackage{booktabs}

\newcommand{\resourcelinks}{%
  \par\smallskip
  {\small
  \setlength{\tabcolsep}{2pt}%
  \begin{tabular}{@{}rl@{}}
    \raisebox{-0.15em}{\includegraphics[height=1.1em]{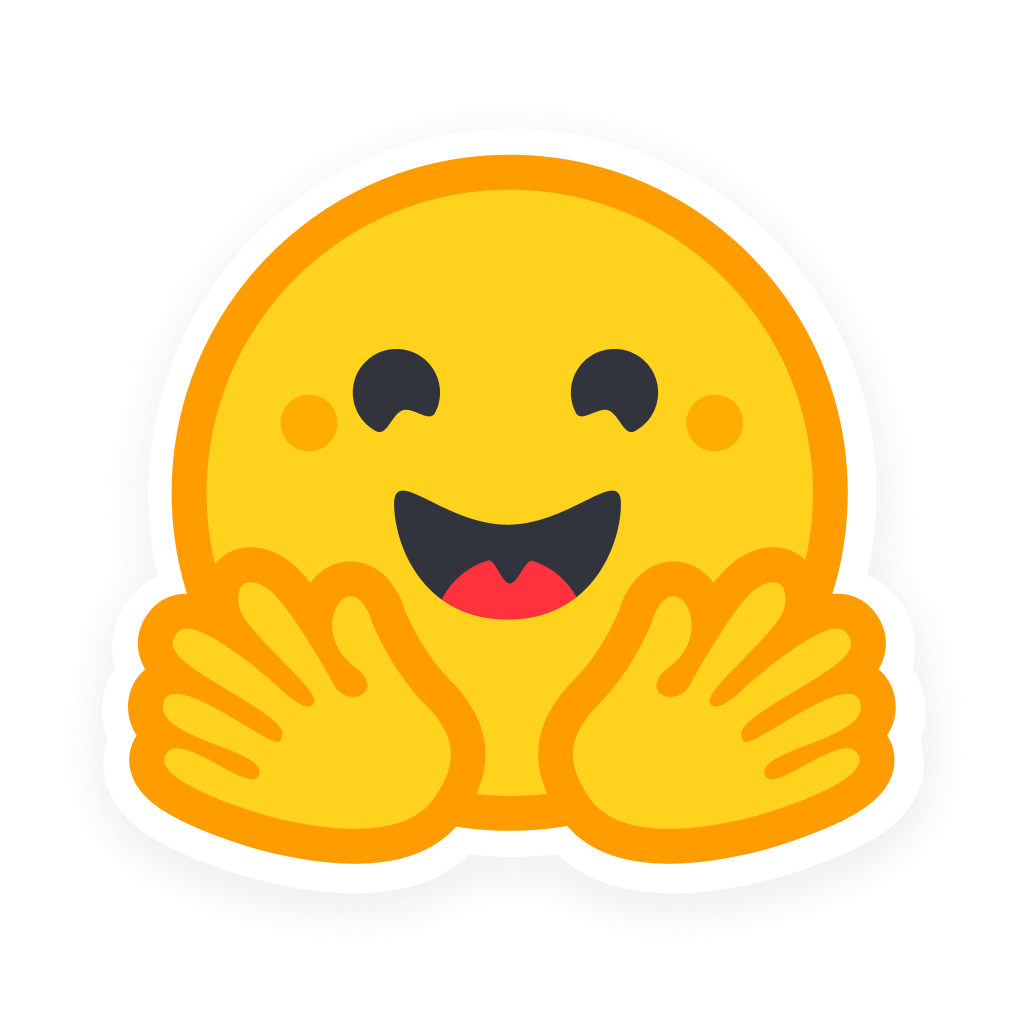}} &
    \textbf{Dataset: }
    \href{https://huggingface.co/datasets/llm-jp/WAON}{llm-jp/WAON}
    \\[0.3em]
    \raisebox{-0.15em}{\includegraphics[height=1.1em]{assets/hf-logo.png}} &
    \textbf{Benchmark: }
    \href{https://huggingface.co/datasets/llm-jp/WAON-Bench}{llm-jp/WAON-Bench}
    \\[0.3em]
    \raisebox{-0.15em}{\includegraphics[height=1.1em]{assets/hf-logo.png}} &
    \textbf{Model:} \href{https://huggingface.co/llm-jp/waon-siglip2-base-patch16-256}{llm-jp/waon-siglip2-base-patch16-256}
    \\[0.3em]
    \raisebox{-0.15em}{\includegraphics[height=1.1em]{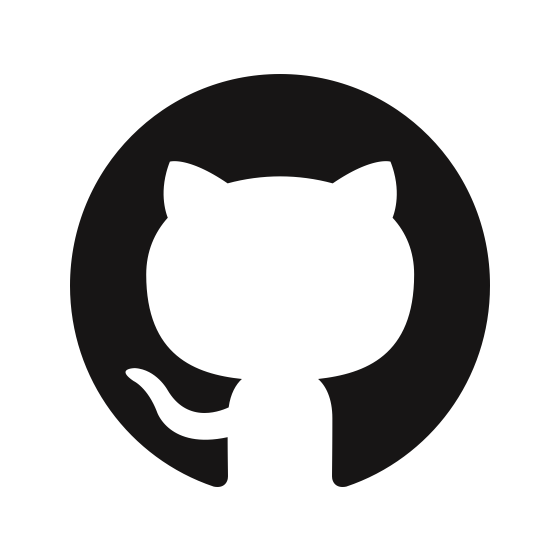}} &
    \textbf{Code: }
    \href{https://github.com/llm-jp/WAON}{github.com/llm-jp/WAON}
  \end{tabular}
  }
  \par
}
\usepackage{multirow}
\usepackage{multicol}
\usepackage[whole]{bxcjkjatype} %
\usepackage{array}
\usepackage[table]{xcolor}

\title{WAON: A Large-Scale Japanese Image-Text Dataset for Cultural Adaptation in Contrastive Vision-Language Models}

\author{Issa Sugiura$^{1,2}$
  Shuhei Kurita$^{3,2}$
   Yusuke Oda$^{2}$
   Daisuke Kawahara$^{4,2}$\\
   {\bf Yasuo Okabe$^{5}$
  Naoaki Okazaki$^{1,2,6}$} \\
  $^1$Institute of Science Tokyo
  $^2$NII LLMC
  $^3$NII
  $^4$Waseda University
  $^5$Kyoto University
  $^6$AIST
}

\begin{document}
\maketitle
\begin{abstract}
Contrastive vision-language models have achieved remarkable progress through large-scale pretraining. Recent work has shown that removing English-only caption filters and pretraining on global data is effective for improving multicultural performance.
We study whether such global pretraining is sufficient for culture-specific understanding, or whether further adaptation with natively sourced data can boost performance beyond what global pretraining alone achieves.
To enable this investigation, we present WAON, the largest publicly available native Japanese image-text dataset constructed from native Japanese web content in Common Crawl, containing approximately 155 million examples. We also introduce WAON-Bench, a manually curated Japanese cultural benchmark spanning 374 classes.
Through comparative fine-tuning experiments on multiple Japanese image-text datasets, we observe that models fine-tuned on WAON consistently achieve stronger performance on Japanese cultural benchmarks than those fine-tuned on English-to-Japanese translated data.
Controlled experiments at matched scale, filtering, and training budget across two model families further indicate that native web origin is the primary driver of this gain.
We release our dataset, benchmark, model, and code.
\resourcelinks
\end{abstract}

\begin{figure}[t]
\centering
\includegraphics[width=\linewidth]{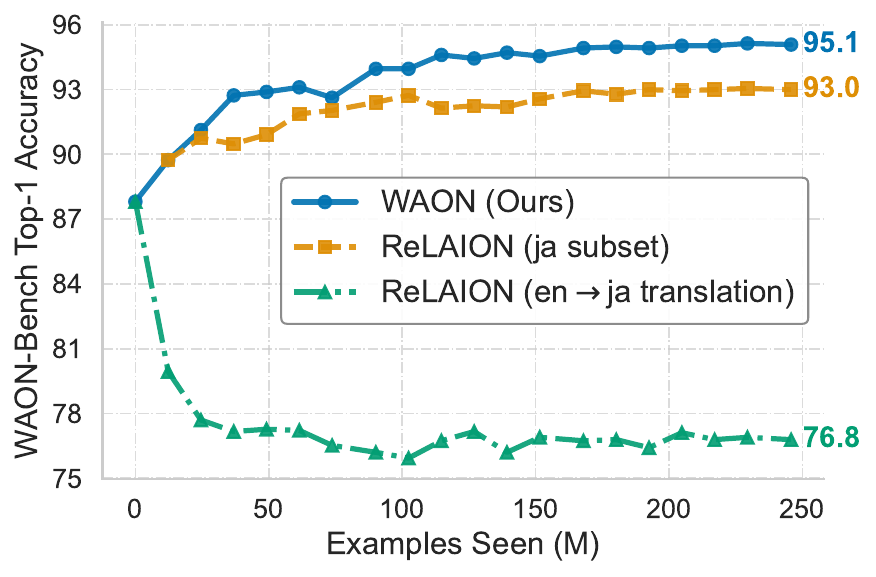}
\caption{Zero-shot top-1 accuracy on WAON-Bench (a Japanese cultural image classification benchmark) over training steps. Fine-tuning on ReLAION (en$\to$ja translation) degrades performance from 87.8\% to 76.8\%, while WAON achieves the highest accuracy of 95.1\%.}
\label{fig:accuracy_curve}
\end{figure}

\begin{table*}[t]
\centering
\setlength{\tabcolsep}{3pt}
\small
\begin{tabular}{lrccc}
\toprule
\textbf{Dataset} & \textbf{Examples} & \textbf{Source} & \textbf{Caption Source} & \textbf{Quality Filter}\\
\midrule
WIT (ja subset){\scriptsize~\citep{Krishna2021wit}} & 1M & Wikipedia & Native & --\\
llm-jp-japanese-image-text-pairs{\scriptsize~\citep{sasagawa-etal-2025-constructing}} & 6.6M & Common Crawl & Native & OpenCLIP + Ja CLIP\\
ReLAION (ja subset){\scriptsize~\citep{schuhmann2022laion}} & 120M & Common Crawl & Native & mCLIP\\
ReLAION (en$\to$ja translation){\scriptsize~\citep{sugiura-etal-2025-developing}} & 1,452M & Common Crawl & Translation & mCLIP\\
\midrule
\textbf{WAON (Ours)} & 155M & Common Crawl & Native & SigLIP2\\
\bottomrule
\end{tabular}
\caption{Comparison of publicly available Japanese image-text pair datasets.
``Native'' indicates that captions are sourced directly from Japanese web
content; ``Translation'' indicates captions obtained by translating those
from English datasets.
WAON is the largest dataset with natively
sourced Japanese captions.}
\label{tab:dataset_comparison}
\end{table*}

\section{Introduction}

Contrastive pre-training on large-scale image-text pair datasets has achieved remarkable progress in learning aligned representations of images and text~\citep{dosovitskiy2021an,radford2021learning,zhai2023sigmoid}. Early work typically filtered training data to English-only captions~\citep{Changpinyo_2021CC12M,radford2021learning,fang2023datafilteringnetworks}, resulting in models heavily biased toward English and Western content. Recent work has shown that removing this filter retains images from diverse cultures and that pretraining on such global data is effective for improving multicultural performance~\citep{globerson2024nofilter,nguyen2024multilingual}, leading to the release of strong open-weight multilingual models such as SigLIP2 and MetaCLIP2~\citep{tschannen2025siglip2multilingualvisionlanguage,chuang2025metaclip2worldwide}.

We study whether such global pretraining is sufficient for culture-specific understanding, or whether further adaptation with natively sourced data can boost performance beyond what global pretraining alone achieves, taking Japanese as a case study.
To enable this investigation, we present \textbf{WAON}\footnote{\textit{Waon} (和音) means ``harmony'' in Japanese.} (\textbf{W}eb-scale image text \textbf{A}ligned \textbf{O}pen \textbf{N}ihongo), the largest publicly available native Japanese image-text dataset constructed from native Japanese web content in Common Crawl, containing 155M examples, filtered using SigLIP2. We also introduce \textbf{WAON-Bench}, a manually curated benchmark for Japanese cultural image classification with 1,870 images across 374 classes spanning 8 diverse categories, covering more than twice as many cultural concepts as the previous benchmark~\citep{honda2024recruit}.

Our experiments reveal that fine-tuning on data translated from English into Japanese consistently degrades Japanese cultural understanding, and WAON achieves the best performance on Japanese cultural benchmarks among publicly available models with comparable architectures. Controlled experiments that match dataset scale, filtering pipeline, and training budget, using both SigLIP2 and OpenAI CLIP as backbones, confirm that this gap stems from native web origin rather than from scale or filtering. These results suggest that translation alone cannot compensate for the absence of natively sourced data.

In summary, our contributions are: (1) we release WAON, the largest publicly available natively sourced Japanese image-text dataset containing 155M examples; (2) we introduce WAON-Bench, a culturally comprehensive Japanese cultural image classification benchmark that more than doubles the class coverage of the prior benchmark; and (3) we provide controlled empirical evidence that native web origin is a critical factor for cultural adaptation in contrastive vision-language models.

\section{Related Work}
\label{sec:related}

\paragraph{Cultural understanding in contrastive vision-language models.}
Early work on contrastive vision-language models, spanning both model development and dataset construction, commonly filtered training data to English-only image-text pairs under the assumption that this would improve data quality~\citep{radford2021learning,schuhmann2022laion,gadre2023datacompsearchgenerationmultimodal,fang2023datafilteringnetworks}.
However, recent work has shown that incorporating non-English and geographically diverse data improves cultural understanding in vision-language models~\citep{globerson2024nofilter,nguyen2024multilingual}.
\citet{nguyen2024multilingual} show that non-English data contains unique visual information and broader cultural concepts not captured by English-only corpora, and that training on translated versions of such data improves both English-centric~\citep{deng2009imagenet} and geographically diverse tasks~\citep{ramaswamy2023geode}. No~Filter~\citep{globerson2024nofilter} highlights the importance of culturally diverse data for training contrastive models.
Meta~CLIP~2~\citep{chuang2025metaclip2worldwide} trains contrastive vision-language models on global data at scale.
These works highlight the importance of incorporating non-English data during training. Building on these findings, our work further demonstrates that, in the setting of Japanese cultural adaptation via fine-tuning vision-language models pretrained on global data, native Japanese data remains crucial: fine-tuning with English-to-Japanese translated data degrades Japanese cultural understanding, whereas natively collected Japanese data more effectively improves performance on Japanese cultural tasks.

\paragraph{Image-text pair datasets.}
Image-text pair datasets have played a central role in advancing contrastive vision-language models~\citep{radford2021learning}. Early efforts relied
on small-scale datasets annotated by human experts~\citep{deng2009imagenet,
lin2015microsoftcococommonobjects}. With the introduction of
CLIP~\citep{radford2021learning}, which leveraged web-scale data, the
construction of large-scale datasets from sources such as Common Crawl became
a major research focus. Among these, LAION-5B~\citep{schuhmann2022laion} has
been the most influential, serving as the first public billion-scale dataset
that has powered a wide range of models~\citep{Cherti2023openclip,
Rombach2022stablediffusion,Lin2024vila}.
Alongside scaling datasets, recent research has prioritized data quality.
DataComp~\citep{gadre2023datacompsearchgenerationmultimodal} introduced a
systematic benchmark for evaluating dataset curation methods.
DFN~\citep{fang2023datafilteringnetworks} developed a data filtering model
that enables training high-performance CLIP models.
Meta~CLIP~\citep{xu2024demystifying} proposed a metadata-based curation
approach that yields higher-quality data than the original CLIP dataset.

\paragraph{Japanese vision-language resources.}
For Japanese, several image-text pair datasets exist as shown in Table~\ref{tab:dataset_comparison}, yet large-scale datasets free from English bias remain scarce. WIT~\citep{Krishna2021wit} is constructed by collecting image-text pairs from Wikipedia, but contains only 1M Japanese examples, insufficient for extensive fine-tuning.
ReLAION (ja subset)~\citep{schuhmann2022laion} is the Japanese subset of ReLAION-5B~\citep{schuhmann2022laion}, a large-scale multilingual image-text dataset, containing 120M examples. However, it relies on mCLIP~\citep{chen-etal-2023-mclip} for quality filtering. mCLIP is trained on a machine-translated extension of CC12M~\citep{Changpinyo_2021CC12M}, which was constructed by filtering web-crawled data to English captions only, and is substantially smaller in scale than recent multilingual models~\citep{tschannen2025siglip2multilingualvisionlanguage,chuang2025metaclip2worldwide}. These limitations introduce English bias~\citep{globerson2024nofilter} and limited performance, potentially causing mCLIP to inadequately score Japanese culture-specific image-text pairs during filtering.
ReLAION (en$\to$ja translation)~\citep{sugiura-etal-2025-developing} is constructed by translating the English captions of the ReLAION English subset into Japanese, covering 1,452M image-text pairs, but introduces translation errors and inherits cultural biases from the source data.
Evaluation resources are equally limited: the only existing Japanese cultural benchmark, Recruit~\citep{honda2024recruit}, comprises 161 classes heavily concentrated in food categories, offering limited coverage of Japanese culture as a whole. WAON and WAON-Bench address both gaps simultaneously.

\section{Building WAON}

We construct \textbf{WAON}, a large-scale Japanese image-text pair dataset
based on Common Crawl~\citep{commoncrawl2025commoncrawl}. We first extract
(image URL, caption) pairs from Japanese HTML pages in Common Crawl snapshots,
and then apply filtering and deduplication to obtain 155M Japanese image-text
pairs. We describe the construction pipeline below.

\begin{figure}[t]
\centering
\includegraphics[width=0.85\linewidth]{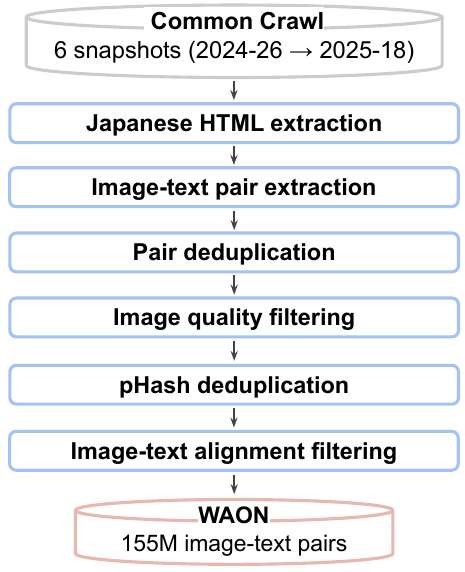}
\caption{Construction pipeline of WAON. Beginning with six Common Crawl snapshots, we extract Japanese image-text pairs and apply cascaded filtering and deduplication, yielding 155M high-quality pairs.}
\label{fig:construction_pipeline}
\end{figure}

\subsection{Dataset Construction Pipeline}
\label{sec:construction_pipeline}
Figure~\ref{fig:construction_pipeline} illustrates our dataset construction
pipeline. We first extract Japanese HTML documents from WARC files as
efficiently as possible to reduce the number of samples processed in later,
computationally intensive steps such as model-based filtering. After extracting
image-text pairs from the HTML, we perform text-based deduplication before
downloading images, followed by image-based filtering and deduplication. We
apply this pipeline to six recent Common Crawl snapshots (2024-26, 2024-33,
2024-42, 2024-51, 2025-08, and 2025-18), iteratively refining each step
through manual inspection of sampled subsets following established
practices~\citep{schuhmann2022laion,lee2022deduplicating,zhu2023multimodal}.

\paragraph{Japanese HTML extraction.}
Common Crawl provides archives in three formats: WARC, WAT, and WET. We use
WARC files, which contain complete HTML documents and corresponding request
metadata. Each snapshot contains approximately 90,000 to 100,000 WARC files,
each containing multiple raw HTML documents.

We retain only Japanese HTML documents, as they are more likely to contain
Japanese captions and culturally relevant images~\citep{nguyen2024multilingual}. Following
\citet{okazaki2024building}, we first determine language based on the
\texttt{lang} attribute in the HTML header, and apply model-based language
identification only to documents that pass this initial filter, as running it
on all HTML bodies would be computationally prohibitive at our scale. We then
apply Trafilatura~\citep{barbaresi-2021-trafilatura} to extract the underlying
text, and Lingua~\citep{stahl2022lingua} to confirm the document language. We
also discard documents with an empty \texttt{<title>} tag, as high-quality
image-text pairs are more likely to appear in well-structured HTML pages.

\paragraph{Image-text pair extraction.}
From the extracted HTML, we obtain (image URL, caption) pairs. Captions are
collected from either the \texttt{alt} attribute of the image tag or the
corresponding \texttt{<figcaption>} element. We remove entries with invalid
image URLs and those whose captions contain no Japanese characters (i.e., no
Unicode code points in the Hiragana, Katakana, or CJK Unified Ideographs
ranges).

\paragraph{Pair deduplication.}
Duplicates on the web often comprise low-diversity content such as advertising
images, logos, and simple graphics. Following prior
work~\citep{lee2022deduplicating,zhu2023multimodal}, we deduplicate on both
image URLs and captions, retaining only the first occurrence of each (image
URL, caption) pair. To improve memory efficiency, we use a Bloom filter, a
probabilistic data structure that may introduce false positives but guarantees
no false negatives.

\begin{figure}[t]
  \centering
\includegraphics[width=\linewidth]{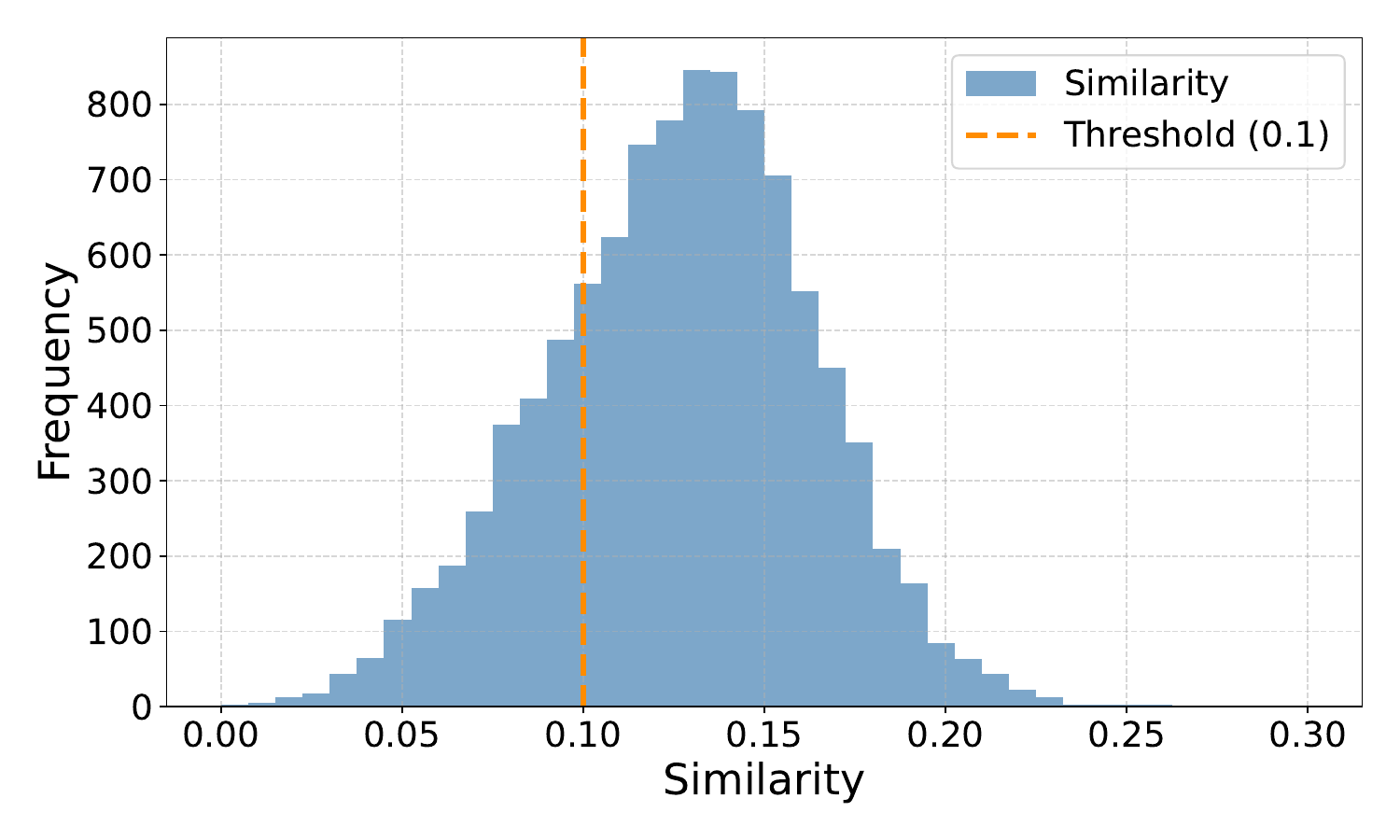}
\caption{Distribution of SigLIP2-base cosine similarity scores for 10,000 randomly sampled image-text pairs prior to alignment filtering. Pairs below the threshold of 0.1 are removed.}
\label{fig:siglip_score}
\end{figure}

\begin{table}[t]
\centering
\small
\begin{tabular}{cr}
\toprule
\textbf{Snapshot} & \textbf{Examples}\\
\midrule
CC-MAIN-2025-18 & 37,445,634 \\
CC-MAIN-2025-08 & 36,043,758  \\
CC-MAIN-2024-51 & 28,178,004 \\
CC-MAIN-2024-42 & 20,221,965 \\
CC-MAIN-2024-33 & 17,910,213 \\
CC-MAIN-2024-26 & 15,433,133 \\
\midrule
Total & 155,232,707 \\
\bottomrule
\end{tabular}
\caption{Number of image-text pairs per Common Crawl snapshot in WAON after
cross-snapshot deduplication of image URLs, captions, and pHash values.
Earlier snapshots contain a higher proportion of previously seen content,
resulting in fewer remaining pairs after deduplication.}
\label{tab:num_deduplicated}
\end{table}

\paragraph{Image quality filtering.}
We download images from the deduplicated URL list using
\texttt{img2dataset}~\citep{beaumont-2021-img2dataset}, a tool designed for
parallel downloading of web images. We then apply heuristic filtering to remove
low-quality content: following mmC4~\citep{zhu2023multimodal}, we discard
images whose width or height is below 150 pixels or whose aspect ratio falls
outside $[0.5, 2.0]$, as abnormal aspect ratios are indicative of
advertisements or banners.

To remove unsafe content, we apply \texttt{dataset2metadata}~\citep{
gadre2023datacompsearchgenerationmultimodal}, an NSFW classifier based on
OpenCLIP~\citep{Cherti2023openclip}. We set the filtering threshold to 0.1,
determined by analyzing the score distribution and manually inspecting randomly
sampled images across score ranges, following prior
work~\citep{zhu2023multimodal,sasagawa-etal-2025-constructing}. This process
effectively removes most photographic and comic adult content.

\paragraph{pHash deduplication.}
Even after the previous steps, many near-duplicate images remain that are
visually similar despite having different URLs and captions. We compute
perceptual hashes (pHash) using ImageHash~\citep{buchner2021imagehash} and
deduplicate using a Bloom filter. We opt for exact matching rather than
Hamming distance-based deduplication, as the latter requires computing
pairwise distances across all images, which is computationally prohibitive
at our scale.

\begin{figure*}[t]
  \centering
  \includegraphics[width=\linewidth]{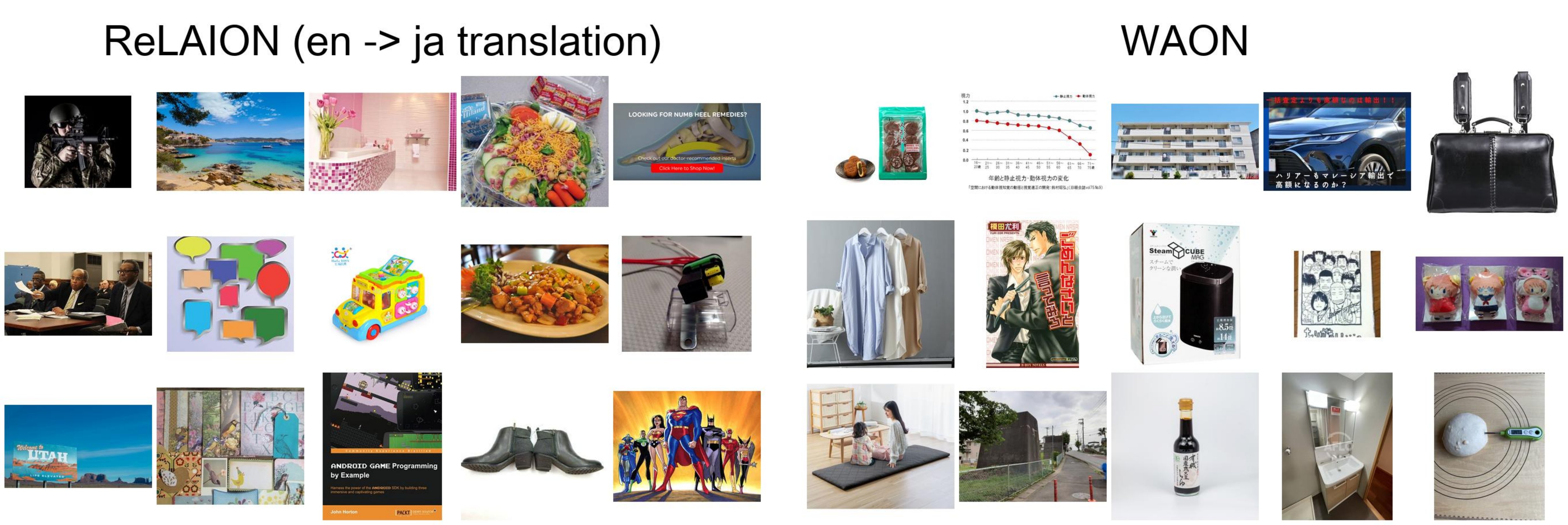}
  \caption{Randomly sampled images from WAON and ReLAION (en$\to$ja translation)~\citep{sugiura-etal-2025-developing}.
  WAON contains many images with Japanese text, scenes captured in Japan, and culturally Japanese content, whereas ReLAION (en$\to$ja) more frequently contains images associated with English text and Western cultural contexts.}
  \label{fig:relaion_vs_waon}
\end{figure*}
\paragraph{Image-text alignment filtering.}
Many image-text pairs on the web exhibit poor semantic alignment between
visual and textual content, as captions are created during HTML authoring
without strict consistency requirements. Following LAION~\citep{schuhmann2022laion},
we filter misaligned pairs using cosine similarity between image and text
embeddings computed by SigLIP2
(\texttt{siglip2-base-patch16-256})~\citep{tschannen2025siglip2multilingualvisionlanguage}.
We use SigLIP2 instead of mCLIP~\citep{chen-etal-2023-mclip}, which was used in prior work~\citep{schuhmann2022laion}, because SigLIP2 is trained on the multilingual WebLI dataset~\citep{chen2023pali} containing 12B image-text pairs across 109 languages, whereas mCLIP relies on translation-based multilingual transfer from a substantially smaller dataset.
In our preliminary evaluations, SigLIP2 exhibited strong performance on Japanese cultural understanding tasks, making it better suited for evaluating semantic alignment in Japanese image-text pairs.
Figure~\ref{fig:siglip_score} shows the score distribution for 10,000 randomly
sampled pairs from the CC-MAIN-2025-18 snapshot prior to filtering. Based on
manual inspection of 100 randomly sampled pairs across score ranges, we set
the threshold to 0.1 and discard pairs below this value, as pairs above this
threshold were generally well aligned semantically.

\paragraph{Cross-Snapshot Deduplication and Statistics.}
We run the pipeline described above independently for each snapshot, then perform cross-snapshot deduplication of image URLs, captions, and pHash values in reverse chronological order from CC-MAIN-2025-18 to CC-MAIN-2024-26. As earlier snapshots contain a higher proportion of previously seen content, more pairs are removed during deduplication, resulting in fewer remaining examples (Table~\ref{tab:num_deduplicated}). In total, WAON contains approximately 155M image-text pairs.

\subsection{Qualitative Analysis}

To qualitatively examine the characteristics of WAON, we randomly sample image-text pairs from WAON and compare them with samples from ReLAION (en$\to$ja translation)~\citep{sugiura-etal-2025-developing}. Figure~\ref{fig:relaion_vs_waon} shows 15 randomly sampled images from each dataset.
WAON contains more images with Japanese text, scenes photographed in Japan, and content closely related to Japanese culture. In contrast, ReLAION (en$\to$ja) more frequently includes images containing English text and content associated with Western cultural contexts. These observations suggest that the two datasets follow notably different distributions, and that WAON contains a larger proportion of Japanese-associated visual content.\footnote{We also present additional representative examples from WAON in Appendix~\ref{sec:waon_examples}.}

\begin{figure*}[t]
\begin{center}
\includegraphics[width=\linewidth]{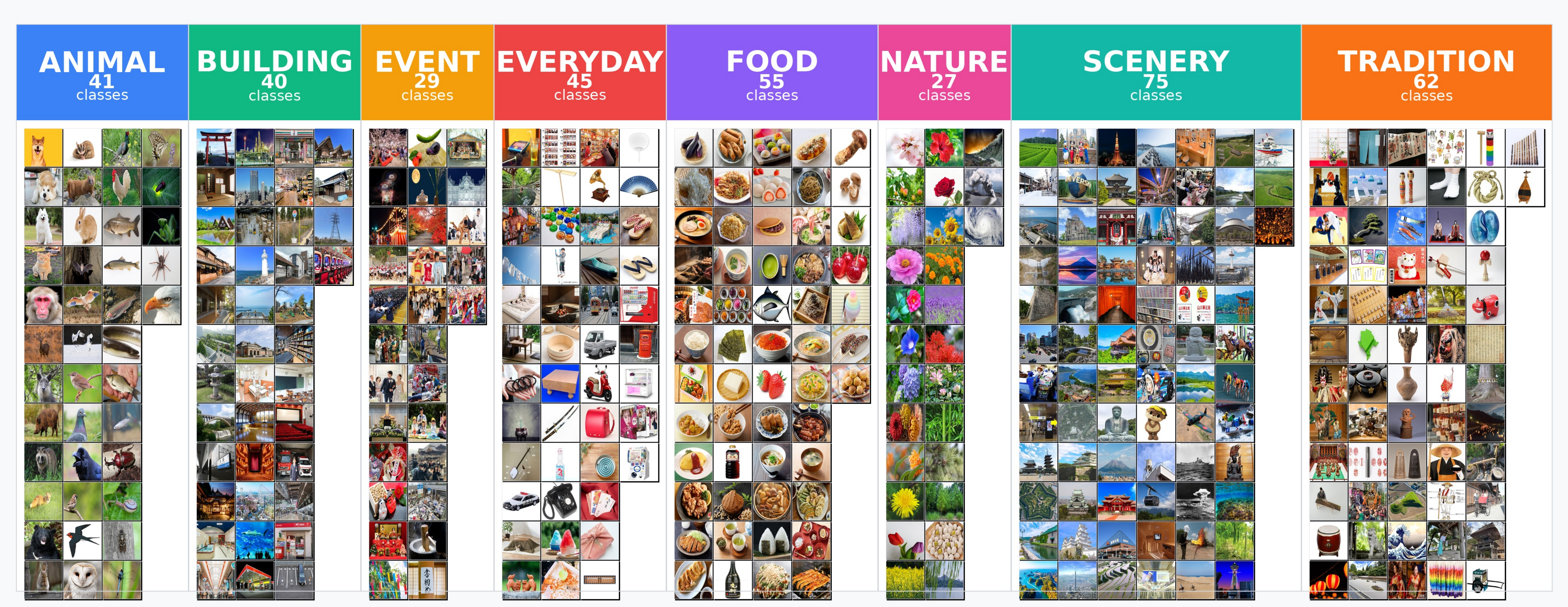}
\caption{Overview of WAON-Bench. WAON-Bench comprises 374 classes spanning 8 categories (animal, building, event, everyday, food, nature, scenery, and tradition), with one randomly sampled image shown for each class.}
\label{fig:waon-bench}
\end{center}
\end{figure*}

\section{Building WAON-Bench}
The Recruit dataset~\citep{honda2024recruit} has been widely used to evaluate Japanese cultural understanding in contrastive vision-language models~\citep{clip-japanese-base}, but its limited cultural coverage of only 161 classes heavily skewed toward food leaves many aspects of Japanese culture underrepresented.

To address this limitation, we construct WAON-Bench, a manually
curated Japanese cultural image classification benchmark designed to address
the limited cultural coverage of the Recruit dataset~\citep{honda2024recruit},
which contains only 161 classes heavily skewed toward food. As illustrated in
Figure~\ref{fig:waon-bench}, WAON-Bench comprises 374 classes spanning
8 categories (animal, building, event, everyday, food, nature, scenery, and
tradition), with 5 images per class for a total of 1,870 images, providing
more than twice the cultural coverage of the previous benchmark with a more
balanced distribution across diverse cultural domains.

\subsection{Dataset Construction Pipeline}

\paragraph{Category definition.}
We define eight categories to broadly cover diverse concepts in Japanese
culture: animal, building, event, everyday, food, nature, scenery, and
tradition.

\paragraph{Class definition.}
We collect class names related to Japanese culture from diverse sources,
including existing datasets~\citep{honda2024recruit}, Japanese Wikipedia
articles, Google searches, conversations with LLMs, and direct observation
during visits to Japanese cities such as Kyoto, resulting in 374 class names
(e.g., Shiba Inu, Tokyo Tower).\footnote{See Appendix~\ref{appendix:waonbench_classes} for the complete class list.} We iteratively add
classes while maintaining approximate balance across categories. Although the
correspondence between categories and classes is inherently ambiguous, we
assign each class to a single category for organizational convenience.

\paragraph{Image collection.}
We collect images for each class by using the class name as a search query in Google Image Search and manually selecting five images from the results.
We limit collection to five images per class because, as observed in the Recruit dataset~\citep{honda2024recruit}, retrieving more images frequently introduces noisy examples that do not match the class label, and for many culturally specific classes, only a limited number of clearly relevant images could be reliably identified. While this limits within-class visual diversity, expanding image counts via in-the-wild photography collection remains an important direction for future work. During selection, we prioritize diversity in composition, perspective, and setting, and exclude images containing objects corresponding to other classes to avoid label ambiguity.

\paragraph{Dataset verification.}
To ensure data quality, we recruit two native Japanese-speaking undergraduate student annotators and ask them to verify the correspondence between each image and its class label.
We ask the annotators to check not only whether the image matches the assigned label but also whether it contains elements corresponding to other class labels.\footnote{See Appendix~\ref{sec:waon-bench-verification-tool} for the annotation interface.}
Each annotator reviews half of the samples, identifying a total of 16 labeling errors, which we subsequently correct.

\subsection{Visual Diversity of WAON-Bench}
To quantitatively evaluate the visual diversity of WAON-Bench, we compare it
with the Recruit dataset. We randomly sample 1,000 images from each dataset,
compute image embeddings using the vision encoder of
\texttt{siglip2-base-patch16-256}, and measure the average pairwise cosine
distance after L2 normalization. WAON-Bench achieves 0.495, higher than
Recruit (0.425), indicating that WAON-Bench images are more widely dispersed
in the embedding space and thus more visually diverse.

\begin{figure}[t]
\centering
\includegraphics[width=\linewidth]{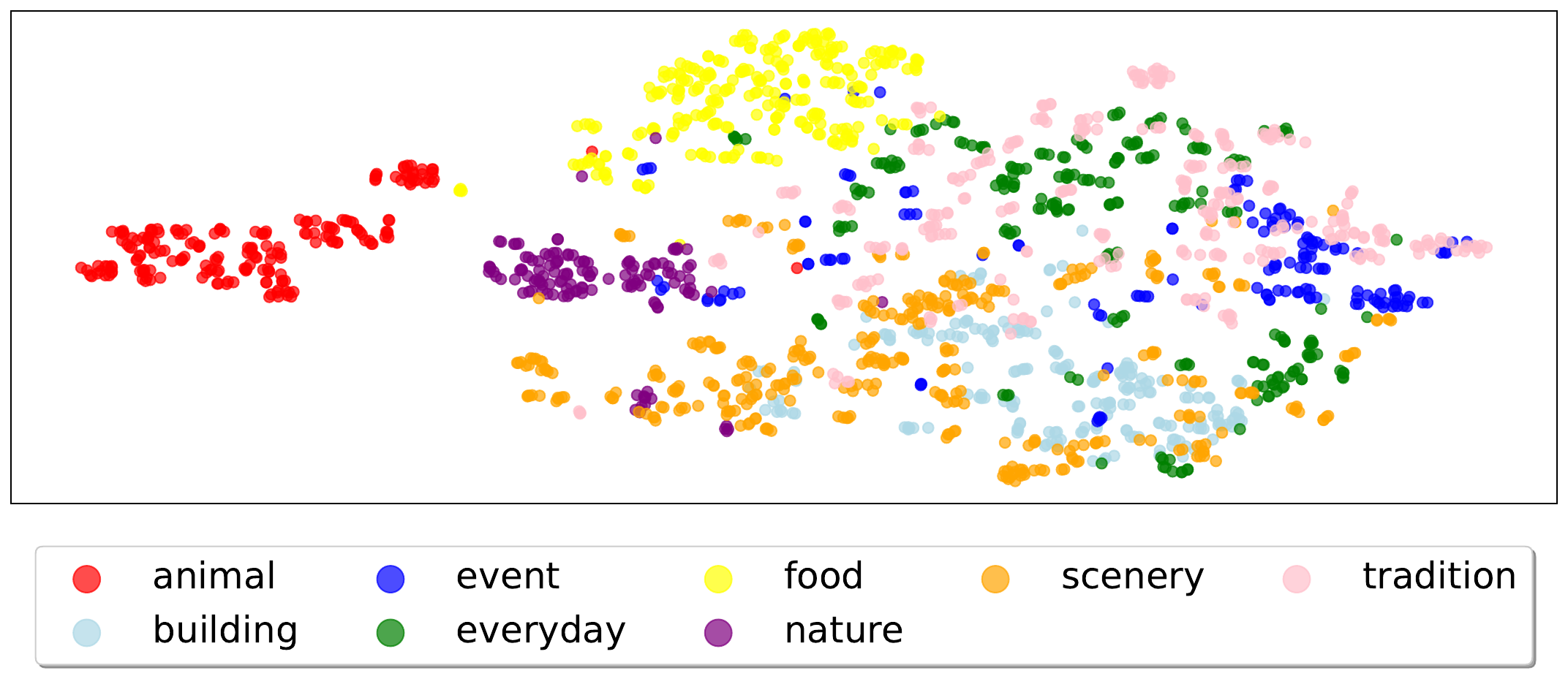}
\caption{t-SNE visualization of image embeddings for WAON-Bench images, colored by
category. The animal, food, and nature categories form relatively distinct
clusters, while the remaining categories are more intermixed, reflecting the
inherent visual overlap between cultural concepts that span multiple domains.}
\label{fig:waon-bench-tsne}
\end{figure}

To further examine the semantic distribution of WAON-Bench, we apply t-SNE
to the same embeddings (Figure~\ref{fig:waon-bench-tsne}). The animal, nature,
and food categories form relatively distinct clusters, while the remaining
categories are more intermixed, reflecting the inherent visual overlap between
cultural concepts that span multiple categories. Since category labels are not
used in our experiments, we do not further refine the category-class mapping
and leave this as future work.

\section{Experiments}
To evaluate the quality of WAON and existing Japanese image-text datasets, we fine-tune \texttt{siglip2-base-patch16-256}~\citep{tschannen2025siglip2multilingualvisionlanguage}, a strong multilingual model pretrained on global data, on WAON and two existing Japanese datasets. We conduct an experiment in which only the fine-tuning dataset is varied, while the model architecture, compute budget, and hyperparameter configuration are kept fixed, ensuring that the training data itself is the primary factor affecting model performance.

\subsection{Training Settings}
\label{sec:training_settings}
We fine-tune each model for 30,000 steps with a batch size of 8,192, totaling
approximately 245M example exposures. We use AdamW~\citep{loshchilov2018decoupled}
with an epsilon of 1e-8, a maximum learning rate of 1e-5 with cosine decay
scheduling, a warmup of 1,500 steps, and a minimum learning rate of 1e-7.
We use the SigLIP loss~\citep{zhai2023sigmoid} as our training objective.
All experiments are conducted on eight NVIDIA H200 GPUs.

\subsection{Training Datasets}
\label{sec:training_datasets}
To evaluate the effectiveness of WAON and existing large-scale Japanese image-text pair datasets as fine-tuning data, we use three Japanese large-scale datasets: WAON, ReLAION (ja subset)~\citep{schuhmann2022laion}, and ReLAION (en$\to$ja translation)~\citep{sugiura-etal-2025-developing}.

WAON is our newly constructed dataset consisting of 155M image-text pairs collected from Common Crawl.
ReLAION (ja subset)~\citep{schuhmann2022laion} is a Japanese subset extracted from the multilingual ReLAION-5B dataset~\citep{schuhmann2022laion} using CLD3 language identification~\citep{ooms2020cld3}. It contains 120M examples, of which 85M pairs were successfully downloaded as of June 2025 due to broken image URLs. This dataset's quality filtering relies on mCLIP~\citep{chen-etal-2023-mclip}.
ReLAION (en$\to$ja translation)~\citep{sugiura-etal-2025-developing} is a Japanese-translated version derived from the English subset of ReLAION-5B. It was constructed by translating the original English captions into Japanese using Gemma 2~\citep{gemmateam2024gemma2improvingopen}, resulting in 1,452M image-text pairs.

To prevent data contamination, we decontaminate all training datasets at the image level against every evaluation set, including WAON-Bench: we compute pHash values for all evaluation images and remove any training image whose pHash matches, so that no benchmark image appears in the training data.
We exclude smaller datasets such as WiT~\citep{Krishna2021wit} and
llm-jp-japanese-image-text-pairs~\citep{sasagawa-etal-2025-constructing}, as their substantially smaller scale
would require more training epochs and lead to different training dynamics,
making fair comparison difficult.

\begin{table*}[t]
\setlength{\tabcolsep}{3pt}
\centering
\small
\begin{tabular}{llccccc}
\toprule
\textbf{Model} &  &\multicolumn{3}{c}{\textbf{Zero-Shot Classification}}& \textbf{Retrieval} & \textbf{Avg.}\\
&& \multicolumn{2}{c}{\textbf{Japanese Culture}}  & \textbf{Western Culture} & \textbf{Global} & \\
&  & \textbf{WAON-Bench}  & \textbf{Recruit} & \textbf{ImageNet} & \textbf{XM3600} \\
\midrule
\rowcolor{gray!10}
\textit{Baseline Models} &&&&&& \\
\multicolumn{2}{l}{clip-japanese-base~\citep{clip-japanese-base}} &   90.1 & 81.7  & 48.9 &78.0 & 74.7\\
\multicolumn{2}{l}{siglip-base-patch16-256-mult~\citep{zhai2023sigmoid}}   & 89.3  & 75.1& 53.3 & 43.2 &65.2\\
\multicolumn{2}{l}{siglip2-base-patch16-256~\citep{tschannen2025siglip2multilingualvisionlanguage}}   & 87.8 & 77.0 & 48.1& 38.3 & 62.8\\
\midrule
\rowcolor{gray!10}
\textit{Fine-tuned Models} & Fine-Tuning Data&&&&&\\
\multirow{3}{*}{siglip2-base-patch16-256} & \textbf{WAON (Ours)} & \textbf{95.1}   & \textbf{83.3} & 49.6 & 73.5 & \textbf{75.4}\\
 & ReLAION (ja subset)  & 93.0 & 81.7 & 47.4 & 72.4&73.6\\
 & ReLAION (en$\to$ja translation)& 76.8 & 67.4  & \textbf{51.7} & \textbf{73.8} & 67.4\\
\bottomrule
\end{tabular}
\caption{Comparison of baseline and fine-tuned models on zero-shot classification
and retrieval tasks. Fine-tuning on WAON achieves the best performance on
Japanese cultural benchmarks and the best overall
average, surpassing both baseline models and those fine-tuned on ReLAION
variants. Notably, fine-tuning on ReLAION (en$\to$ja) consistently degrades
performance on Japanese cultural benchmarks relative to the base model.}
\label{tab:evaluation_result}
\end{table*}

\subsection{Evaluation Datasets}
To evaluate Japanese, Western, and broader global cultural understanding, we use four datasets: WAON-Bench, Recruit~\citep{honda2024recruit}, ImageNet~\citep{deng2009imagenet}, and XM3600~\citep{ThapliyalCrossmodal2022}.

WAON-Bench is our newly constructed Japanese cultural image classification benchmark, consisting of 374 classes and 1,870 examples. Recruit dataset~\citep{honda2024recruit} is also a Japanese cultural image classification benchmark, containing 161 classes and 7,654 examples.
ImageNet is a large-scale image classification benchmark that contains many images associated with Western culture, comprising 1,000 classes and 50,000 examples. We use the Japanese-translated class labels provided by \citet{sawada2024japaneseclip}.
XM3600~\citep{ThapliyalCrossmodal2022} is a global image-caption retrieval benchmark consisting of 3,600 images collected from regions associated with 36 languages. Each image is paired with captions in all 36 languages. We use the Japanese captions in our experiments.

We report top-1 accuracy for classification tasks and Recall@1 for retrieval tasks.

\subsection{Baseline Models}
To provide broader comparison beyond our fine-tuned models, we compare against
publicly released models that use ViT-B/16~\citep{dosovitskiy2021an} as the
vision encoder. Specifically, we include
\texttt{clip-japanese-base}~\citep{clip-japanese-base} as a Japanese-specific
CLIP model, and \texttt{siglip-base-patch16-256-mult}~\citep{zhai2023sigmoid}
and \texttt{siglip2-base-patch16-256}~\citep{
tschannen2025siglip2multilingualvisionlanguage} as strong multilingual
baselines.

\section{Results}
\label{sec:results}
Table~\ref{tab:evaluation_result} reports the performance of both baseline and fine-tuned models across all benchmarks. Figure~\ref{fig:accuracy_curve} shows the accuracy curves on WAON-Bench during fine-tuning with each training dataset.

\paragraph{Translated data degrades Japanese cultural understanding.}
As shown in Figure~\ref{fig:accuracy_curve}, fine-tuning on ReLAION (en$\to$ja translation) consistently degrades Japanese cultural understanding, with WAON-Bench performance steadily declining throughout fine-tuning from 87.8\% to 76.8\%. Furthermore, Table~\ref{tab:evaluation_result} shows that the fine-tuned model underperforms the base model on both WAON-Bench and the Recruit dataset.

In contrast, ReLAION (en$\to$ja translation) achieves the largest improvement among the three training datasets on ImageNet (ja), with a gain of +3.6 points, suggesting that translated data remains effective for improving performance on benchmarks more closely aligned with Western cultural distributions. These results indicate that translated datasets inherit cultural biases from English source data and are not well suited for Japanese culture-specific fine-tuning despite their large scale.

\paragraph{Natively collected data yields superior cultural understanding.}

Fine-tuning on WAON yields consistent improvements on Japanese cultural benchmarks, reaching 95.1\% on WAON-Bench and 83.3\% on Recruit, outperforming all baseline models.
ReLAION (ja subset), also natively sourced, shows gains on Japanese cultural benchmarks, yet lags behind WAON on both WAON-Bench and Recruit.
One possible reason is that its filtering pipeline relies on mCLIP~\citep{chen-etal-2023-mclip}, which is trained on small-scale English data~\citep{Changpinyo_2021CC12M} and may be less effective at preserving some Japanese-specific image-text pairs during filtering. This suggests that stronger multilingual models such as SigLIP2, trained on large-scale global data, could yield more culturally robust filtering and further improve data quality.

Notably, although \texttt{clip-japanese-base} is trained from scratch with heavy pretraining on a large-scale in-house Japanese dataset comprising approximately 540M pairs~\citep{clip-japanese-base}, our model achieves superior performance through fine-tuning on top of a globally pretrained backbone~\citep{tschannen2025siglip2multilingualvisionlanguage}, supporting prior findings~\citep{globerson2024nofilter} that global pretraining followed by culture-specific fine-tuning is an effective strategy.

\begin{table*}[t]
\setlength{\tabcolsep}{3pt}
\centering
\small
\begin{tabular}{llccccc}
\toprule
\textbf{Backbone} & \textbf{Fine-Tuning Data (10M)} &\multicolumn{3}{c}{\textbf{Zero-Shot Classification}}& \textbf{Retrieval} & \textbf{Avg.}\\
&& \multicolumn{2}{c}{\textbf{Japanese Culture}}  & \textbf{Western Culture} & \textbf{Global} & \\
&  & \textbf{WAON-Bench}  & \textbf{Recruit} & \textbf{ImageNet} & \textbf{XM3600} \\
\midrule
\multirow{4}{*}{siglip2-base-patch16-256} & -- (no fine-tuning) & 87.8 & 77.0 & 48.1 & 38.3 & 62.8\\
 & \textbf{WAON (Ours)} & \textbf{91.6} & 79.6 & 47.9 & \textbf{67.6} & \textbf{71.7}\\
 & ReLAION (ja subset) & 90.6 & \textbf{80.1} & 46.3 & 67.1 & 71.0\\
 & ReLAION (en$\to$ja translation) & 85.7 & 72.9 & \textbf{49.5} & 67.2 & 68.8\\
\midrule
\multirow{4}{*}{CLIP ViT-B/16 (OpenAI)} & -- (no fine-tuning) & 12.2 & 10.9 & 3.0 & 1.8 & 7.0\\
 & \textbf{WAON (Ours)} & \textbf{70.2} & \textbf{60.2} & 23.2 & 32.3 & \textbf{46.5}\\
 & ReLAION (ja subset) & 65.2 & 54.8 & 19.7 & 33.9 & 43.4\\
 & ReLAION (en$\to$ja translation) & 40.1 & 35.0 & \textbf{29.0} & \textbf{38.8} & 35.7\\
\bottomrule
\end{tabular}
\caption{Controlled comparison at matched scale, filtering, and training budget: all datasets are re-filtered with WAON's SigLIP2 pipeline, subsampled to 10M pairs, and used to fine-tune each backbone for 10,000 steps. Native corpora outperform the translated corpus on Japanese cultural benchmarks under both backbones, including OpenAI CLIP ViT-B/16, which is not used in WAON's construction. Bold: best fine-tuned result per backbone and column.}
\label{tab:controlled_comparison}
\end{table*}

\begin{table*}[t]
\centering
\small
\setlength{\tabcolsep}{4pt}
\begin{tabular}{llrrr}
\toprule
\textbf{Benchmark} & \textbf{Metric} & \textbf{WAON} & \textbf{ReLAION (ja subset)} & \textbf{ReLAION (en$\to$ja translation)}\\
\midrule
\multirow{2}{*}{WAON-Bench (374 classes)} & Covered classes & 359 (96\%) & 358 (96\%) & 329 (88\%)\\
 & Median mentions & 608 & 820 & 67\\
\midrule
\multirow{2}{*}{Recruit (161 classes)} & Covered classes & 161 (100\%) & 161 (100\%) & 157 (98\%)\\
 & Median mentions & 1,226 & 1,628 & 215\\
\bottomrule
\end{tabular}
\caption{Coverage of WAON-Bench and Recruit class names in the captions of each 10M subset: number of classes mentioned at least once (covered classes) and median number of captions mentioning a class (median mentions). Native corpora mention Japanese cultural concepts 5--12$\times$ more often than the translated corpus.}
\label{tab:cultural_coverage}
\end{table*}

\paragraph{WAON maintains competitive performance on global benchmarks.}
Fine-tuning on WAON also improves performance beyond Japanese cultural tasks. On XM3600 (ja), WAON yields a substantial gain of 35.2 points over the base model (38.3$\to$73.5), approaching the performance of ReLAION (en$\to$ja) (73.8). Increasing the number of training steps may allow WAON to achieve performance comparable to clip-japanese-base (78.0).

\section{Isolating the Effect of Native Origin}
\label{sec:controlled_comparison}
\paragraph{Setup.}
The datasets in Table~\ref{tab:evaluation_result} differ not only in caption origin but also in scale, filtering model, and image distribution, and SigLIP2 serves both as WAON's filter and as the fine-tuned backbone. To remove these confounds, we re-filter ReLAION (ja subset) and ReLAION (en$\to$ja translation) with the same SigLIP2-based alignment filtering used for WAON (Section~\ref{sec:construction_pipeline}), randomly sample a 10M subset from each of the three datasets, and decontaminate them as in Section~\ref{sec:training_datasets}. On each subset, we fine-tune \texttt{siglip2-base-patch16-256} and OpenAI CLIP ViT-B/16~\citep{radford2021learning}, which is not used anywhere in WAON's construction, for 10,000 steps with the hyperparameters of Section~\ref{sec:training_settings}. This equalizes filtering, scale, and training budget, leaving caption origin as the primary variable.

\paragraph{Results.}
Under these controls (Table~\ref{tab:controlled_comparison}), both native corpora clearly outperform the translated corpus on WAON-Bench and Recruit with both backbones, while WAON and ReLAION (ja subset) become comparable. The cultural gain thus stems from native web origin rather than from the SigLIP2 filter, dataset scale, or shared architecture between filter and model.

\paragraph{Cultural-concept coverage.}
\label{sec:cultural_coverage}
To quantify \emph{why} native corpora help, we count, for each WAON-Bench and Recruit class, the captions in each 10M subset containing the Japanese class name (Table~\ref{tab:cultural_coverage}). All corpora cover most classes, but the native corpora contain 5--12$\times$ more mentions per class than the translated corpus. Because the underlying pages and images are predominantly Western, translated captions rarely surface Japanese-specific concepts even when fluent.

\section{Conclusion}
We presented WAON, a 155M-example Japanese image-text dataset built from native Japanese web content, and WAON-Bench, a manually curated Japanese cultural image classification benchmark. Fine-tuning on WAON yields the best performance on Japanese cultural benchmarks, whereas English-to-Japanese translated data degrades it, and controlled experiments at matched scale, filtering, and training budget across two backbones show that native web origin is the primary driver of this gain.

\section*{Limitations}
\paragraph{Japanese as a single case study.}
We take Japanese as a case study, and whether our findings generalize to other languages and cultures remains an open question. Extending this investigation to other low-resource or culturally distinct languages is an important direction for future work.
\paragraph{Model scale and architecture.}
Our main experiments fine-tune \texttt{siglip2-base-patch16-256} only. Although the controlled comparison in Section~\ref{sec:controlled_comparison} confirms that the native-versus-translated ordering also holds for OpenAI CLIP ViT-B/16, both are base-sized dual encoders, and we do not assess whether our findings hold for larger models or different architectural designs.
\paragraph{Scope of downstream evaluation.}
This paper targets the representation quality of contrastive dual-encoder models and evaluates zero-shot classification and image-text retrieval only. Whether WAON also benefits generative tasks such as visual question answering, captioning, or multimodal reasoning, which require a decoder-based model, remains to be evaluated in future work.
\paragraph{Dataset construction pipeline.}
Our construction pipeline includes multiple filtering and deduplication steps, but we do not provide a quantitative analysis of each step's contribution to overall dataset quality. A more systematic analysis of each pipeline component would be valuable for guiding future large-scale vision-language dataset construction.

\section*{Ethical Considerations}
\paragraph{Web data and copyright.}
WAON is constructed from publicly available web data collected via Common
Crawl. We do not redistribute any image files; instead, we release only image
URLs and associated metadata. All images remain hosted by their original
sources, and their use is subject to the respective website terms and
applicable copyright laws. Users of WAON are responsible for ensuring
compliance with the terms of service of the original sources.

\paragraph{Content filtering and risk mitigation.}
To mitigate potential risks, we apply NSFW filtering to remove unsafe content during dataset construction (Section~\ref{sec:construction_pipeline}). We acknowledge that despite this filtering, some harmful, biased, or privacy-sensitive content may remain, as is inherent in large-scale web-crawled datasets.
We provide clear instructions for ethical use and will maintain a post-release mechanism for reporting and removing problematic samples, including those involving copyright infringement, personal privacy, sensitive content, or cultural bias. We encourage users to exercise caution and report any concerns to the dataset maintainers.

\section*{Acknowledgements}
In this research work, we used the ``mdx: a platform for building data-empowered society''.
We used ABCI 3.0 provided by AIST and AIST Solutions with support from ``ABCI 3.0 Development Acceleration Use''.
This research was supported by the Science Tokyo Support Program for Doctoral Students, funded by the Universities for International Research Excellence.

\bibliography{custom}

\appendix

\section{Licenses for Our Resources}
WAON is released under the Apache 2.0 License. Note that we distribute only image URLs rather than the images themselves. WAON-Bench, our fine-tuned SigLIP2 model, and our code are also released under the Apache 2.0 License.

\section{Use of AI Assistants}
We used AI assistants to correct typographical errors, improve the clarity and naturalness of expressions, and generate scripts for plotting figures.

\section{Representative Examples in WAON}
\label{sec:waon_examples}

Table~\ref{tab:waon_examples} shows representative examples from WAON, selected to highlight the cultural diversity of the dataset. The examples
illustrate that WAON covers a wide variety of Japan-specific content, including
event flyers, location maps, and food imagery, reflecting the breadth of
Japanese web content captured in the dataset.

\begin{table*}[t]
\small
\centering
\begin{tabular}{p{0.1\linewidth}p{0.24\linewidth}p{0.25\linewidth}p{0.27\linewidth}}

\toprule
\textbf{Image} &
\includegraphics[width=\linewidth,height=3cm,keepaspectratio]{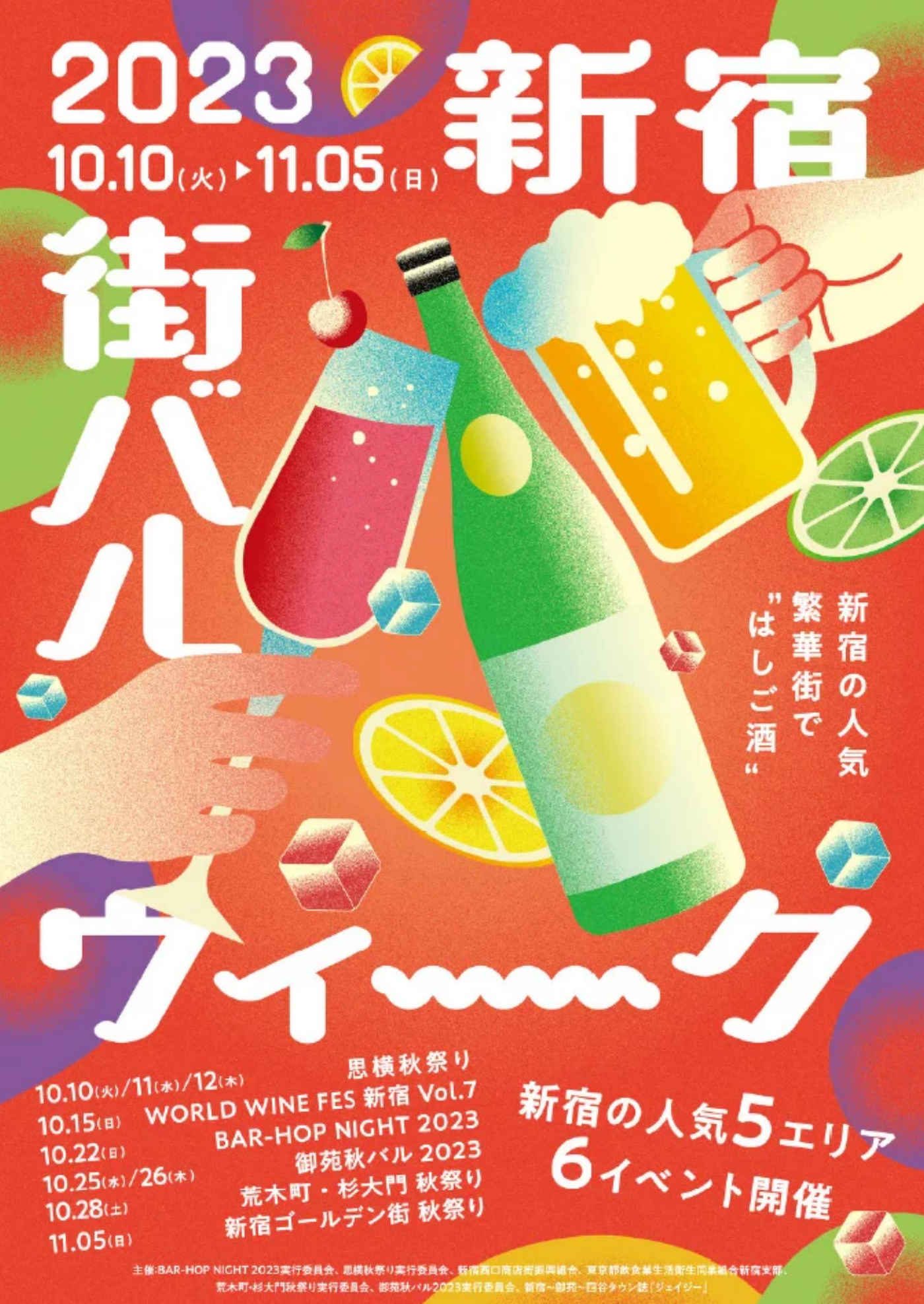} &
\includegraphics[width=\linewidth,height=3cm,keepaspectratio]{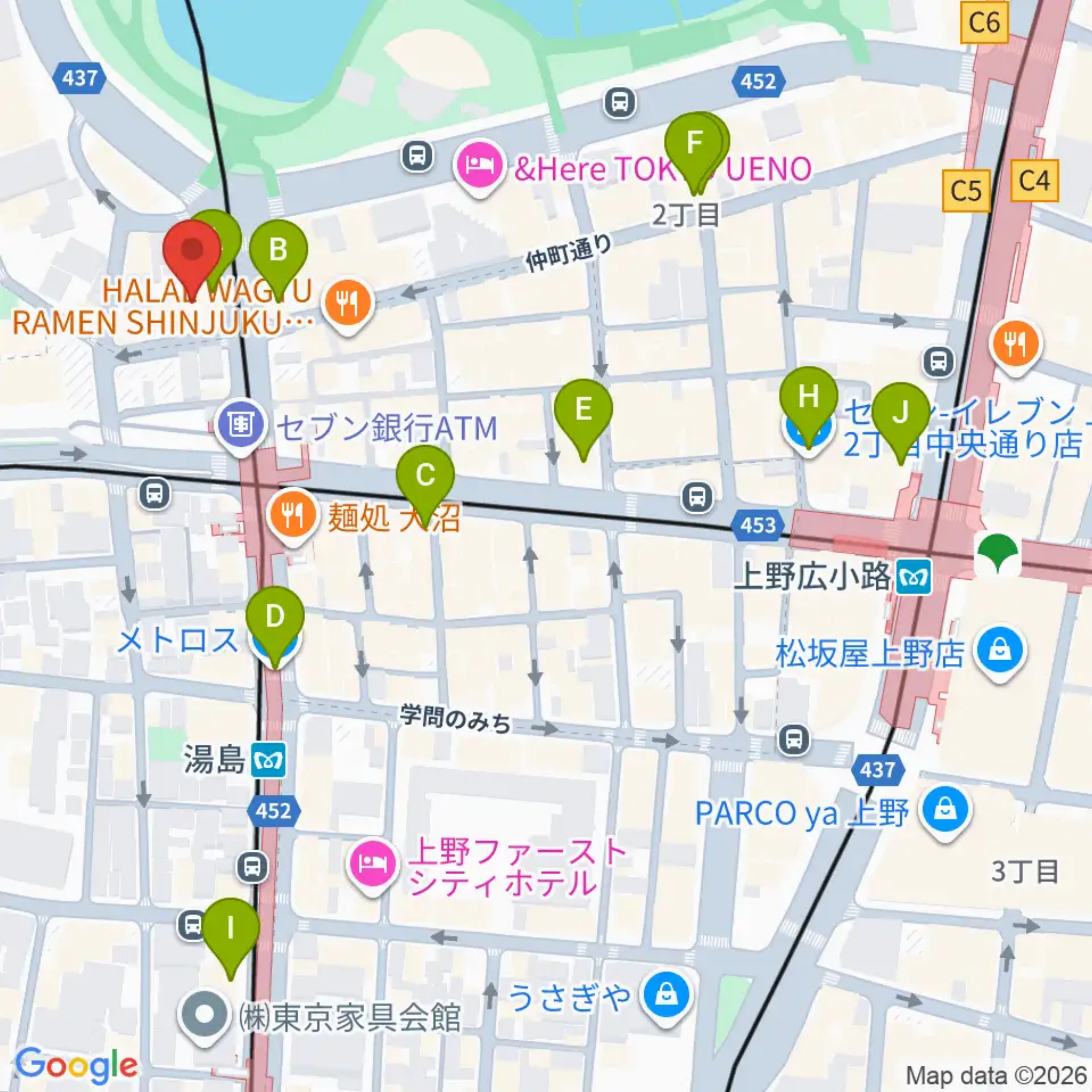} &
\includegraphics[width=\linewidth,height=3cm,keepaspectratio]{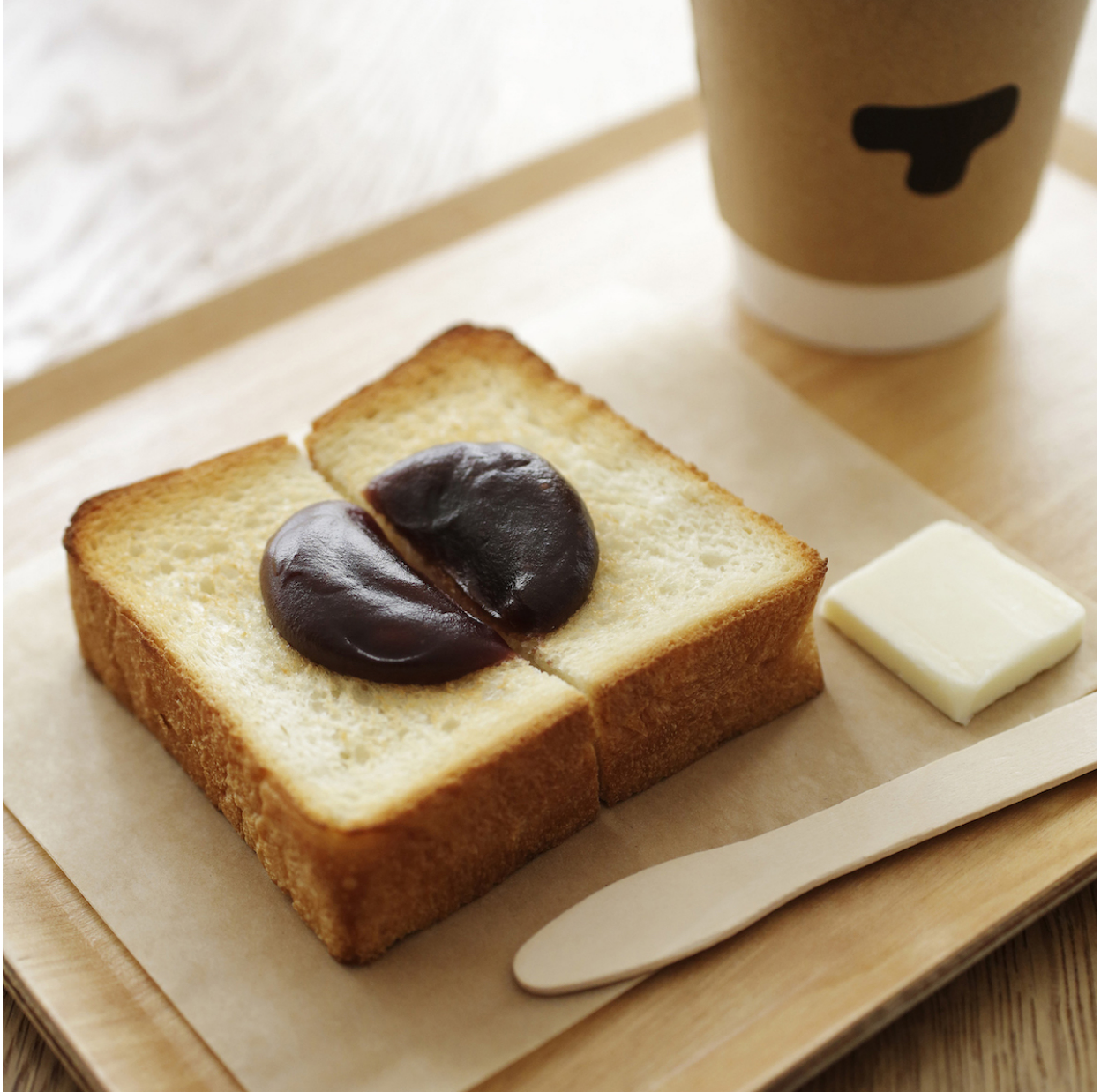} \\
\midrule
\textbf{Caption} & 4年ぶりの開催となる「新宿街バルウィーク2023」
& パールフルートギャラリー東京周辺のコンビニエンスストア一覧地図  & こしあん、黒砂糖とメープルシロップの2種のあんペーストと、[...]\\
&{\scriptsize(``Shinjuku Machi Bar Week 2023,'' held for the first time in four years)} & {\scriptsize(Map of convenience stores around Pearl Flute Gallery Tokyo.)} & {\scriptsize(《An Toast》—a set featuring two types of sweet bean paste (koshian; and a blend of brown sugar and maple syrup),[...])}\\
\midrule
\textbf{Similarity} & 0.154
& 0.165&0.182 \\
\bottomrule
\end{tabular}
\caption{Representative examples of image-text pairs from WAON, covering event
announcements, location maps, and food imagery, reflecting the cultural
diversity of native Japanese web content.}
\label{tab:waon_examples}
\end{table*}

\section{WAON-Bench Class Names}
\label{appendix:waonbench_classes}
Table~\ref{tab:waonbench_classes} lists the class names for each category in WAON-Bench.

\begin{table*}[t]
\centering
\small
\caption{Class names for each category in WAON-Bench.}
\label{tab:waonbench_classes}
\begin{tabular}{lcp{0.72\linewidth}}
\toprule
\textbf{Category} & \textbf{Classes} & \textbf{Class Names} \\
\midrule
Animal& 41 & 'アユ', 'イタチ', 'イノシシ', 'ウグイス', 'ウサギ', 'ウナギ', 'エゾシカ', 'カタツムリ', 'カブトムシ', 'カマキリ', 'カラス', 'キジ', 'キツネ', 'クモ', 'コイ', 'コウモリ', 'スズメ', 'セミ', 'タヌキ', 'チョウ', 'ツキノワグマ', 'ツバメ', 'ツル', 'トキ', 'トンボ', 'ニジマス', 'ニホンカモシカ', 'ニホンザル', 'ニホンリス', 'ニワトリ', 'ハト', 'フクロウ', 'フナ', 'ホタル', 'メダカ', 'モモンガ', '日本スピッツ', '日本猫', '柴犬', '秋田犬', '鷲'
\\
\midrule
Building &40 & 'ガソリンスタンド', 'コンビニ', 'スーパーマーケット', 'バス停', 'パチンコ屋', '交番', '体育館', '倉庫', '公園', '劇場', '博物館', '合掌造り', '商店街', '図書館', '学校', '屋上庭園', '展望台', '工場', '旅館', '映画館', '橋', '水族館', '消防署', '灯台', '灯籠', '町家', '病院', '空港', '納屋', '縁側', '茶室', '遊園地', '郵便局', '鉄塔', '銭湯', '駅', '駐車場', '高層ビル', '高床倉庫', '鳥居'\\
\midrule
Event &29&  'お正月', 'お盆', 'きつねの嫁入り行列', 'ひな祭り', 'よさこい祭り', '七五三', '七夕', '修学旅行', '卒業式', '同窓会', '地域清掃', '地蔵盆', '展示会', '成人式', '書き初め', '月見', '盆踊り', '相撲大会', '節分', '紅葉', '結婚式', '花火大会', '花見', '葬式', '運動会', '遠足', '防災訓練', '雪まつり', '餅つき'\\
\midrule
Everyday &45& 'うちわ', 'おはじき', 'お年玉', 'かき氷', 'こたつ', 'そろばん', 'ちゃぶ台', 'カラオケ', 'ガチャガチャ', 'クレーンゲーム', 'シーサー', 'トミカ', 'プリクラ機', 'ランドセル', '三味線', '下駄', '原付バイク', '囲炉裏', '囲碁盤', '射的', '屋台', '市民プール', '扇子', '数珠', '敷布団', '新幹線', '日本刀', '洗濯物', '温泉', '瓶ラムネ', '竹とんぼ', '竹馬', '線香', '自動販売機', '花札', '草履', '蓄音機', '蚊取り線香', '路面電車', '軽トラ', '郵便ポスト', '金魚すくい', '風呂敷', '風呂桶', '黒電話' \\
\midrule
Food&55& 'いくら', 'いちご', 'いなり寿司', 'うどん', 'うなぎの蒲焼', 'おでん', 'おにぎり', 'お好み焼き', 'きんぴらごぼう', 'しゃぶしゃぶ', 'しらす', 'すき焼き', 'そば', 'たけのこ', 'たこ焼き', 'ちゃんぽん', 'ちらし寿司', 'とんかつ', 'どら焼き', 'ひつまぶし', 'りんご飴', 'わたあめ', 'オムライス', 'カレーライス', 'チャーハン', 'チョコバナナ', 'ベビーカステラ', 'マグロ', 'ラーメン', '味噌', '味噌汁', '和菓子', '唐揚げ', '大福', '天ぷら', '弁当', '抹茶', '日本酒', '春巻き', '松茸', '椎茸', '海苔', '漬物', '焼きそば', '焼き鳥', '焼肉', '白米', '精進料理', '納豆', '緑茶', '茄子', '茶碗蒸し', '豆腐', '醤油', '餃子'\\
\midrule
Nature&27& 'つばき', 'アサガオ', 'アジサイ', 'キク', 'キンモクセイ', 'タンポポ', 'チューリップ', 'ハイビスカス', 'バラ', 'ヒマワリ', 'マリーゴールド', 'ラベンダー', '台風', '噴火', '彼岸花', '杉', '松', '柳', '桜', '梅', '津波', '牡丹', '竹', '菜の花', '蓮', '藤', '銀杏'\\
\midrule
Scenery&75& 'ご当地マンホール', 'カプセルホテルの内部', 'ゴーヤー畑', 'サロベツ湿原', 'スーパーの試食コーナー', 'ビルの屋上給水タンク', 'プラモデル専門店', 'メイドカフェ', 'ユニバーサルスタジオジャパン', 'ランタンフェスティバル', 'ロープウェイ', 'ワカメ干し', '不動明王像', '五稜郭', '京都タワー', '伏見稲荷大社', '兼六園', '原爆', '原爆ドーム', '厳島神社', '名古屋城', '地下鉄改札口', '地元ゆるキャラ', '地蔵', '大阪城', '天橋立', '姫路城', '富士山', '小樽運河', '三内丸山遺跡', '山火事防止の看板', '幕張メッセ', '平等院鳳凰堂', '戦艦大和', '東京スカイツリー', '東京タワー', '東京ディズニーランド', '東大寺', '松島', '桜島', '棚田', '法隆寺', '流氷観光船', '浅草寺', '清水寺', '漁港', '火炎放射器', '牛丼チェーン店のカウンター席', '特急列車の車内販売', '痛車', '皇居', '知床', '石垣', '石垣島のサンゴ礁', '砂防ダム', '秋葉原の電気街', '競艇', '競輪', '競馬', '茶畑', '蜂の巣駆除', '通天閣', '道の駅', '金閣寺', '鎌倉大仏', '防波堤', '雪国の街並み', '零戦', '電車の吊り広告', '首都高のジャンクション', '首里城', '駅前ロータリー', '鳥取砂丘', '鴨川', '黒部ダム'\\
\midrule
Tradition &62& 'お遍路さん', 'かるた', 'けん玉', 'こけし', 'ししおどし', 'しめ縄', 'だるま落とし', 'ねぶた', 'のれん', '修行僧', '前方後円墳', '剣道', '勾玉', '千羽鶴', '印鑑', '埴輪', '太鼓', '妖怪', '将棋', '尺八', '弓道', '弥生土器', '御朱印', '御神木', '折り紙', '招き猫', '提灯', '枕草子', '柔道', '歌舞伎', '浮世絵', '漆器', '灯籠流し', '熊手', '牛車', '狂言', '狛犬', '獅子舞', '琴', '琵琶', '盆栽', '石庭', '破魔矢', '神輿', '空手', '節句人形', '絵馬', '縄文土器', '能舞台', '花魁', '華道', '観光人力車', '赤べこ', '足袋', '銅鐸', '鐘つき', '鐘楼門', '陶器', '雅楽', '風鈴', '鯉のぼり', '鯛車' \\
\bottomrule
\end{tabular}
\end{table*}

\begin{figure*}[t]
\begin{center}
\fbox{%
  \includegraphics[width=0.98\linewidth]{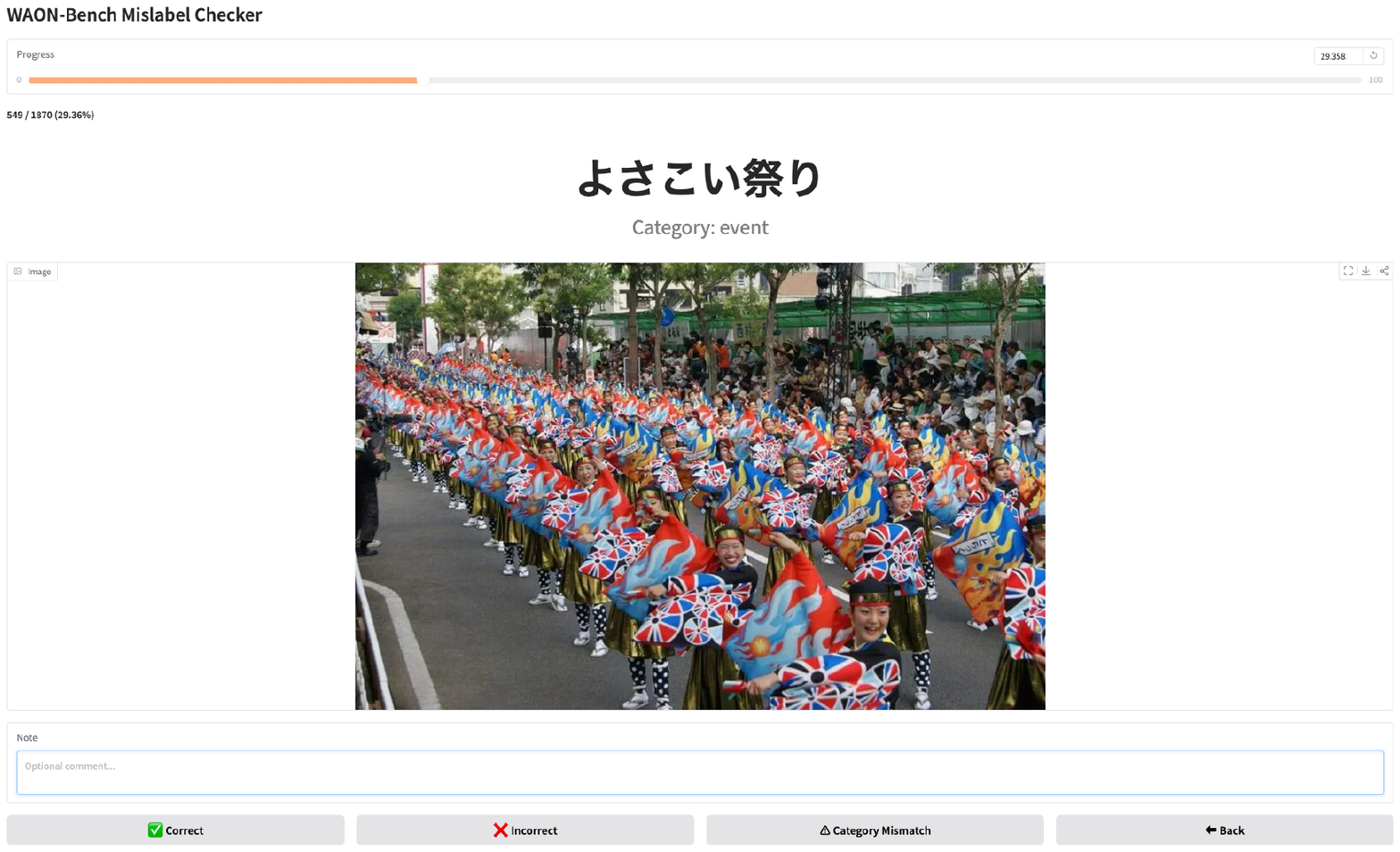}
}
\caption{A screenshot of the annotation interface used to validate the correspondence between classes and images in WAON-Bench.}
\label{fig:annotation_tool}
\end{center}
\end{figure*}

\section{WAON-Bench Verification Tool}
\label{sec:waon-bench-verification-tool}

Figure~\ref{fig:annotation_tool} shows the interface of the verification tool developed using Gradio~\citep{abid2019gradio}, which is used to validate the correspondence between classes and images in WAON-Bench.

\end{document}